\documentclass[9pt,onecolumn,letterpaper]{article}

\usepackage{icb}
\usepackage{times}
\usepackage{epsfig}
\usepackage{graphicx}
\usepackage{amsmath}
\usepackage{amssymb}

\usepackage{subfig}
\usepackage{multirow}
\usepackage{booktabs}

\usepackage[norule]{footmisc}

\usepackage{float} 

\usepackage[breaklinks=true,letterpaper=true,colorlinks,bookmarks=false]{hyperref}

\icbfinalcopy 

\newlength{\w}

\graphicspath{{figures/}}
\DeclareGraphicsExtensions{.eps,.png,.pdf,.jpg,.JPG}

\makeatletter 
\def\ps@IEEEtitlepagestyle{ 
\def\@oddfoot{\mycopyrightnotice} 
\def\@evenfoot{} 
} 
\def\mycopyrightnotice{ 
{\hfill \footnotesize International Conference on Biometrics for Borders
\hfill} 
} 
\makeatother 

\ificbfinal\fi

\begin{document}

\title{Vulnerability of Face Recognition to Deep Morphing}

\author{Pavel Korshunov and
        S\'{e}bastien Marcel\\
Idiap Research Institute, Martigny, Switzerland\\
{\tt\small \{pavel.korshunov,sebastien.marcel\}@idiap.ch}
}

\maketitle
\thispagestyle{empty}

\begin{abstract}

It is increasingly easy to automatically swap faces in images and video or morph two faces into one using generative adversarial networks (GANs). The high quality of the resulted deep-morph raises the question of how vulnerable the current face recognition systems are to such fake images and videos. It also calls for automated ways to detect these GAN-generated faces. In this paper, we present the publicly available dataset of the Deepfake videos with faces morphed with a GAN-based algorithm. To generate these videos, we used open source software based on GANs, and we emphasize that training and blending parameters can significantly impact the quality of the resulted videos. 
We show that the state of the art face recognition systems based on VGG and Facenet neural networks are vulnerable to the deep morph videos, with 85.62\% and 95.00\% false acceptance rates, respectively, which means methods for detecting these videos are necessary.
We consider several baseline approaches for detecting deep morphs and find that the method based on visual quality metrics (often used in presentation attack detection domain) leads to the best performance with 8.97\% equal error rate. Our experiments demonstrate that GAN-generated deep morph videos are challenging for both face recognition systems and existing detection methods, and the further development of deep morphing technologies will make it even more so.
\end{abstract}

{\let\thefootnote\relax\footnotetext{\mycopyrightnotice}}

\section{Introduction}
\label{sec:intro}


Recent advances in automated video and audio editing tools, generative adversarial networks (GANs), and social media allow the creation and the fast dissemination of high quality tampered video content. Such content already led to appearance of deliberate misinformation, coined `fake news', which is impacting political landscapes of several countries~\cite{allcott_gentzkow_2017}. A recent surge of videos (started as obscene) called Deepfakes\footnote{Open source: \url{https://github.com/deepfakes/faceswap}}, in which a neural network is used to train a model to replace faces with a likeness of someone else, are of a great public concern\footnote{{\scriptsize BBC (Feb 3, 2018):} \url{http://www.bbc.com/news/technology-42912529}}. 
Accessible open source software and apps for such face swapping lead to large amounts of synthetically generated Deepfake videos appearing in social media and news, posing a significant technical challenge for detection and filtering of such content. 


Although the original purpose of GAN-based Deepfake is to swap faces of two people in an image or a video, the resulted synthetic face is essentially a morph, i.e., a \textit{deep morph}, of two original faces. The main difference from more traditional morphing techniques is that deep-morph can seamlessly mimic facial expression of the target person and, therefore, can also be successfully used to generate convincing fake videos of people talking and moving about. However, to understand how threatening such videos can be in the context of biometric security, we need to find out whether these deep-morphed videos pose a challenge to face recognition systems and whether they can be easily detected.

\begin{figure*}[tb]
\centering
\subfloat[Morphing faces]{\includegraphics[width=0.4\columnwidth]{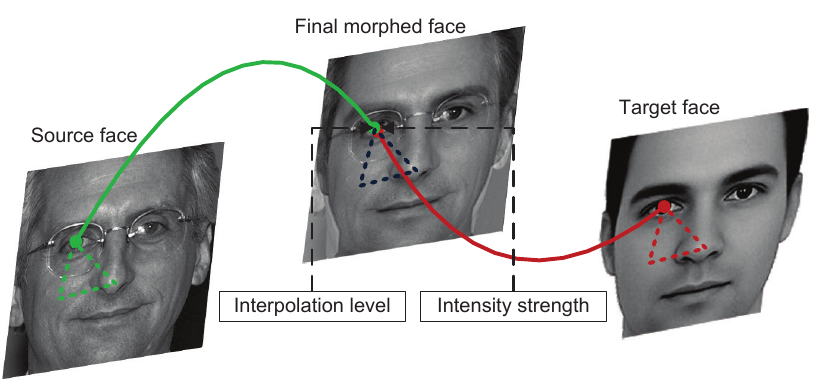}\label{fig:morph}}
\hspace{25pt}
\subfloat[Generating Deepfake faces]{\includegraphics[width=0.3\columnwidth]{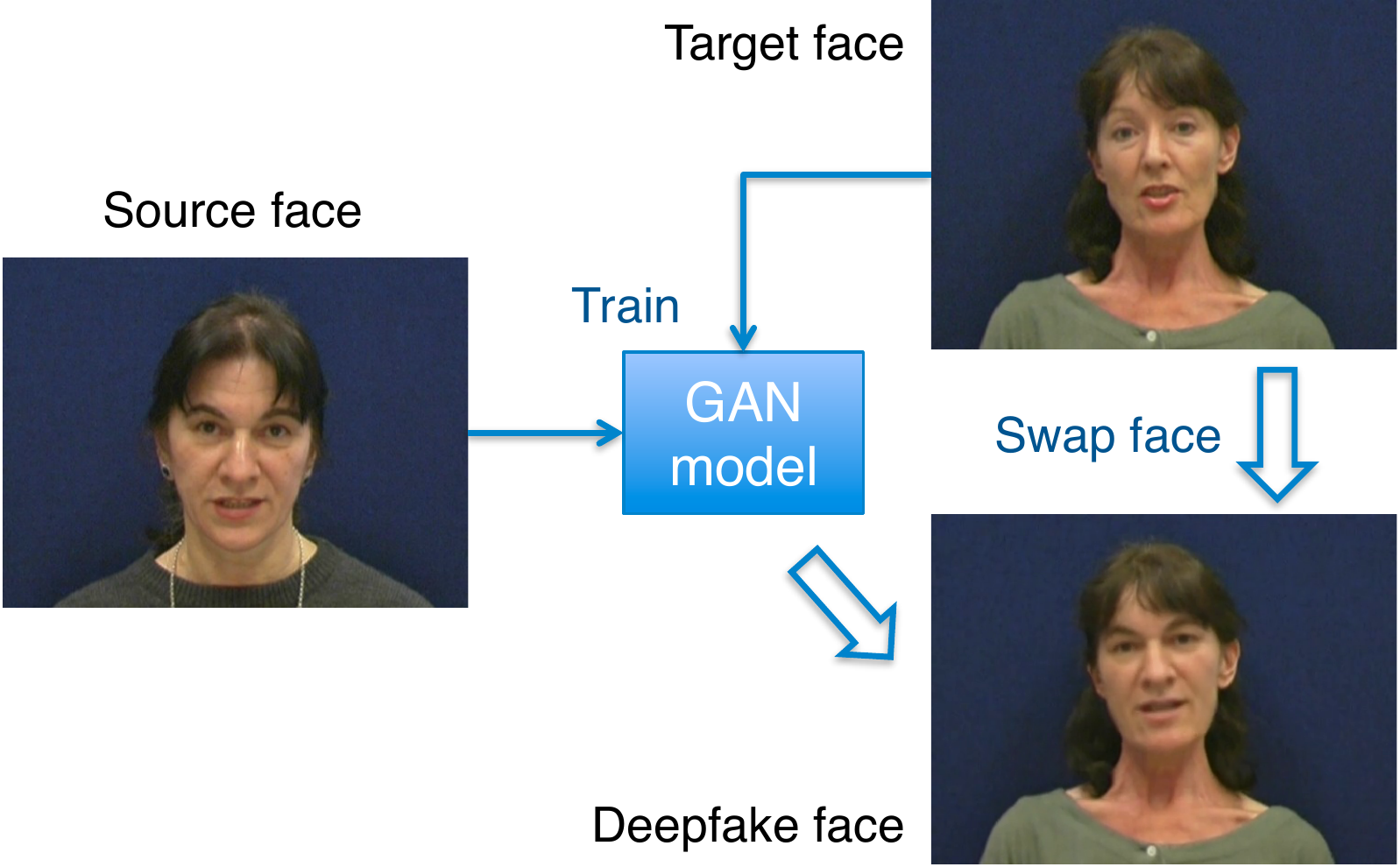}\label{fig:deepfake}}
\caption{Comparing morphing and GAN-based face swapping techniques.}
\label{fig:process}
\end{figure*}

Traditional face morphing (Figure~\ref{fig:morph} illustrates the morphing process) has been shown to be challenging for face recognition systems~\cite{Ferrara2014,Scherhag2019} and several detection methods has been proposed since~\cite{Makrushin2017,Seibold2019,Kramer2019}. For the GAN-based deep-morphing, until recently, most of the research was focusing on advancing the GAN-based face swapping~\cite{Isola2016,Korshunova2017,Nirkin2018,Pham2018}. However, responding to the public demand to detect these synthetic faces, researchers started to work on databases and detection methods, including image and video data~\cite{Verdoliva2018} generated with a previous generation of face swapping approach Face2Face~\cite{Thies2016} or videos collected using Snapchat\footnote{\url{https://www.snapchat.com/}} application~\cite{Agarwal2017}. Several methods for detection of Deepfakes have also been proposed~\cite{Korshunov2019a,Lyu2019,Delp2018}.

\begin{figure*}[tbh]
\centering
\setlength{\w}{0.15\textwidth}
\subfloat{\includegraphics[width=\w]{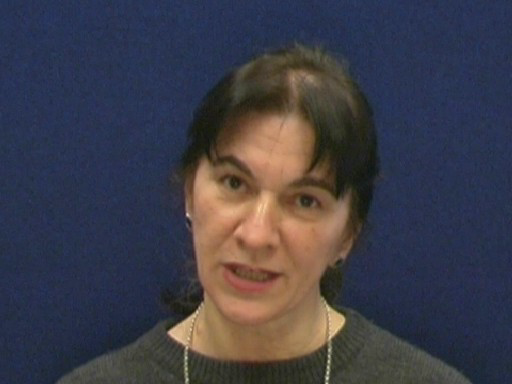}}
\subfloat{\includegraphics[width=\w]{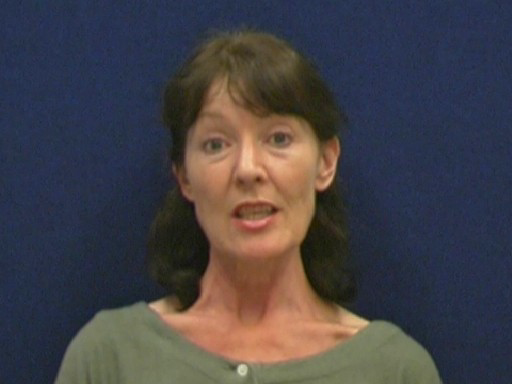}}
\subfloat{\includegraphics[width=\w]{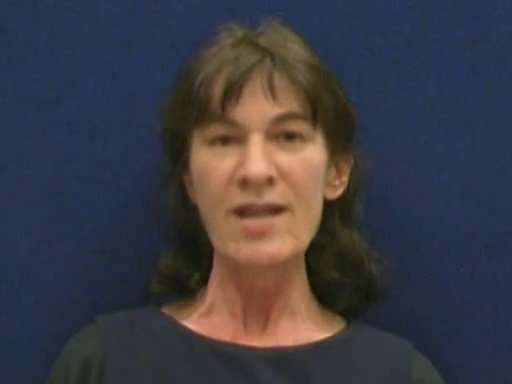}}
\subfloat{\includegraphics[width=\w]{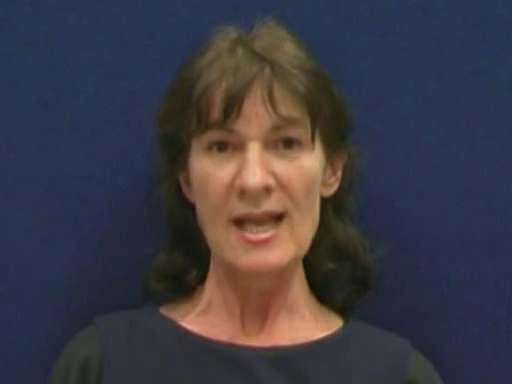}}
\subfloat{\includegraphics[width=\w]{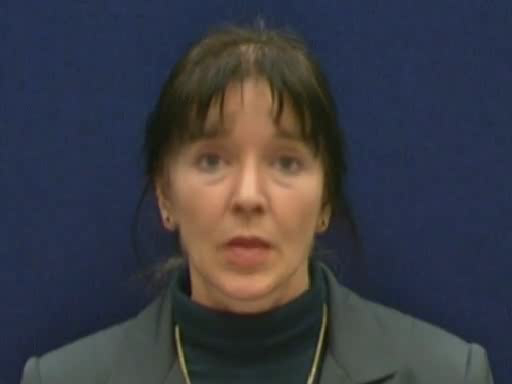}}
\subfloat{\includegraphics[width=\w]{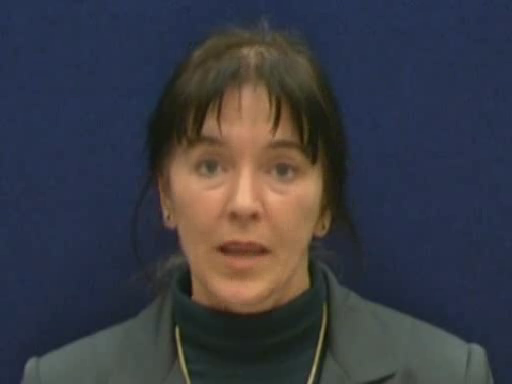}} \\
\subfloat[Original 1]{\includegraphics[width=\w]{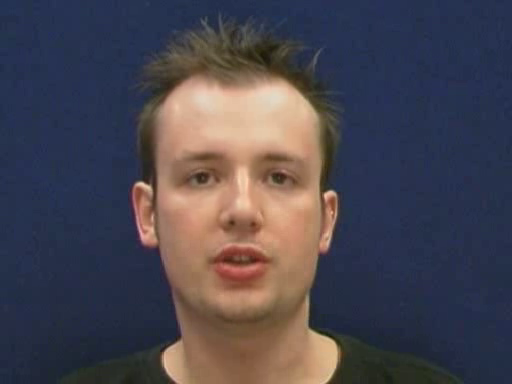}\label{fig:source}}
\subfloat[Original 2]{\includegraphics[width=\w]{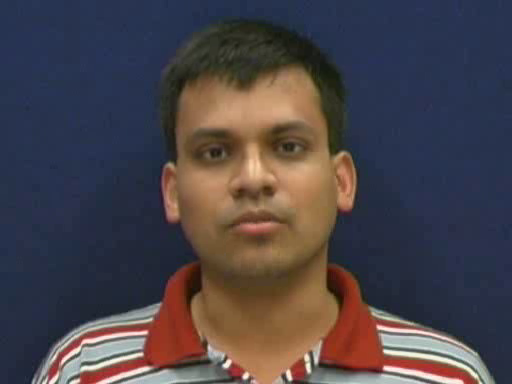}\label{fig:target}}
\subfloat[LQ swap $1\rightarrow 2$]{\includegraphics[width=\w]{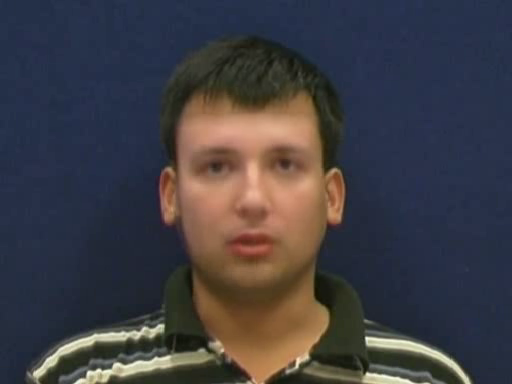}\label{fig:swaplq1}}
\subfloat[HQ swap $1\rightarrow 2$]{\includegraphics[width=\w]{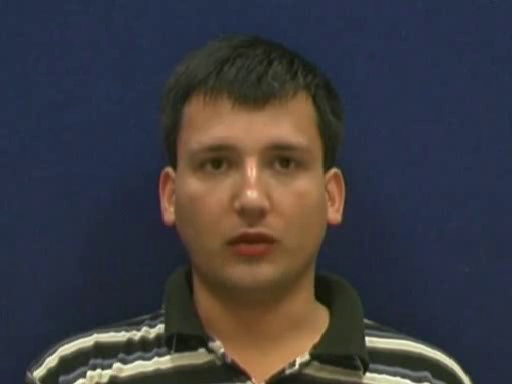}\label{fig:swaphq1}}
\subfloat[LQ swap $2\rightarrow 1$]{\includegraphics[width=\w]{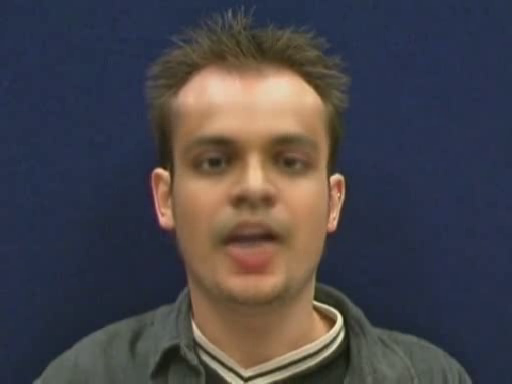}\label{fig:swaplq2}}
\subfloat[HQ swap $2\rightarrow 1$]{\includegraphics[width=\w]{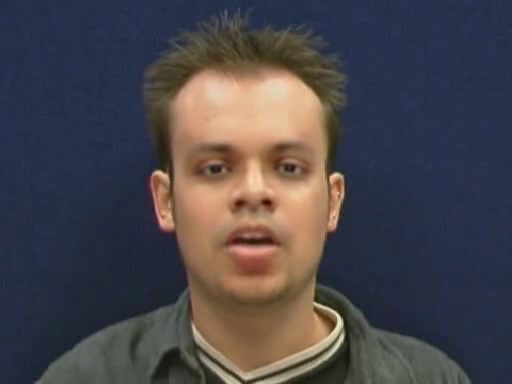}\label{fig:swaphq2}} \\
  
\caption{Screenshot of the original videos from VidTIMIT database and low (LQ) and high quality (HQ) deep morphs.}
\label{fig:examples}
\end{figure*}

In this paper, we focus on evaluating the vulnerability of face recognition systems to Deepfake videos where real faces are replaced by GAN-generated images trained on the faces of two people. 
The resulted synthetic face is essentially a deep morph of two people.
The database was created using the open source software with cyclic GAN model\footnote{\url{https://github.com/shaoanlu/faceswap-GAN}} (see Figure~\ref{fig:deepfake} for illustration), 
which is developed from the original autoencoder-based Deepfake algorithm\footnotemark[1]. We manually selected $16$ similar looking pairs of people from publicly available VidTIMIT database\footnote{\url{http://conradsanderson.id.au/vidtimit/}}. 
For each of $32$ subjects, we trained two different models (see Figure~\ref{fig:examples} for examples), referred to in the paper as the low quality (LQ) model, with $64 \times 64$ input/output size, and the high quality (HQ) model, with $128 \times 128$ size. Since there are $10$ videos per person in VidTIMIT database, we generated $320$ videos corresponding to each version, resulting in total $620$ videos with faces swapped. For the audio, we kept the original audio track of each video, i.e., no manipulation was done to the audio channel. 

We assess the vulnerability of face recognition to deep morph videos using two state of the art systems: based on VGG~\cite{Parkhi15} and Facenet\footnote{\url{https://github.com/davidsandberg/facenet}}~\cite{Schroff2015} neural networks. For detection of the deep morphs, we applied several baseline methods from presentation attack detection domain, by treating deep morph videos as digital presentation attacks~\cite{Agarwal2017}, including simple principal component analysis (PCA) and linear discriminant analysis (LDA) approaches, and the approach based on image quality metrics (IQM) and support vector machine (SVM)~\cite{Galbally2014,Wen2015}.

To allow researchers to verify, reproduce, and extend our work, we provide the database 
coined DeepfakeTIMIT of Deepfake videos\footnote{\url{https://www.idiap.ch/dataset/deepfaketimit}},
face recognition and deep morph detection systems with corresponding scores as an open source Python package 
\footnote{Source code: 
\url{https://gitlab.idiap.ch/bob/bob.report.deepfakes}}.

%
%


\section{Database of deep morph videos}
\label{sec:dataset}

As the original data, we took video from VidTIMIT database\footnotemark[5].
The database contains $10$ videos for each of $43$ subjects, which were shot in controlled environment with people facing camera and reciting predetermined short phrases. From these $43$ subject, we manually selected $16$ pairs in such a way that subjects in the same pair have similar prominent visual features, e.g., mustaches or hair styles. Using GAN-based algorithm based on the available code\footnotemark[4], for each pair of subjects, we generated videos where their faces are replaced by a GAN-generated deep morphs (see the example screenshots in Figure~\ref{fig:examples}). 

For each pair of subjects, we have trained two different GAN models and generated two versions of the deep morphs:
\vspace{-2pt}
\begin{enumerate}
\item {The low quality (LQ) model has input and output image (facial regions only) of size $64 \times 64$. About $200$ frames from the videos of each subject were used for training and the frames were extracted at $4$ fps from the original videos. The training was done for $10'000$ iterations and took about $4$ hours per model on Tesla P40 GPU.}
\vspace{-2pt}

\item{The high quality (HQ) model has input/output image size of $128 \times 128$. About $400$ frames extracted at $8$ fps from videos were used for training, which was done for $20'000$ iterations (about $12$ hours on Tesla P40 GPU).}
\end{enumerate}
\vspace{-2pt}

Also, different blending techniques were used when generating deep morph videos using different models. With LQ model, for each frame from an input video, generator of the GAN model was applied on the face region to generate the fake counterpart. Then a facial mask was detected using a CNN-based face segmentation algorithm proposed in~\cite{Nirkin2018}. Using this mask, the generated fake face was blended with the face in the target video. For HQ model, the blending was done based on facial landmarks (detected with publicly available MTCNN model~\cite{Zhang2016}) alignment between generated fake face and the original face in the target video. Finally, histogram normalization was applied to the blended result to adjust for the lighting conditions, which makes the result more realistic (see Figure~\ref{fig:examples}).

\begin{figure*}[tbh]
\centering
\setlength{\w}{0.35\textwidth}
\subfloat[VGG-based face recognition]{\includegraphics[width=\w]{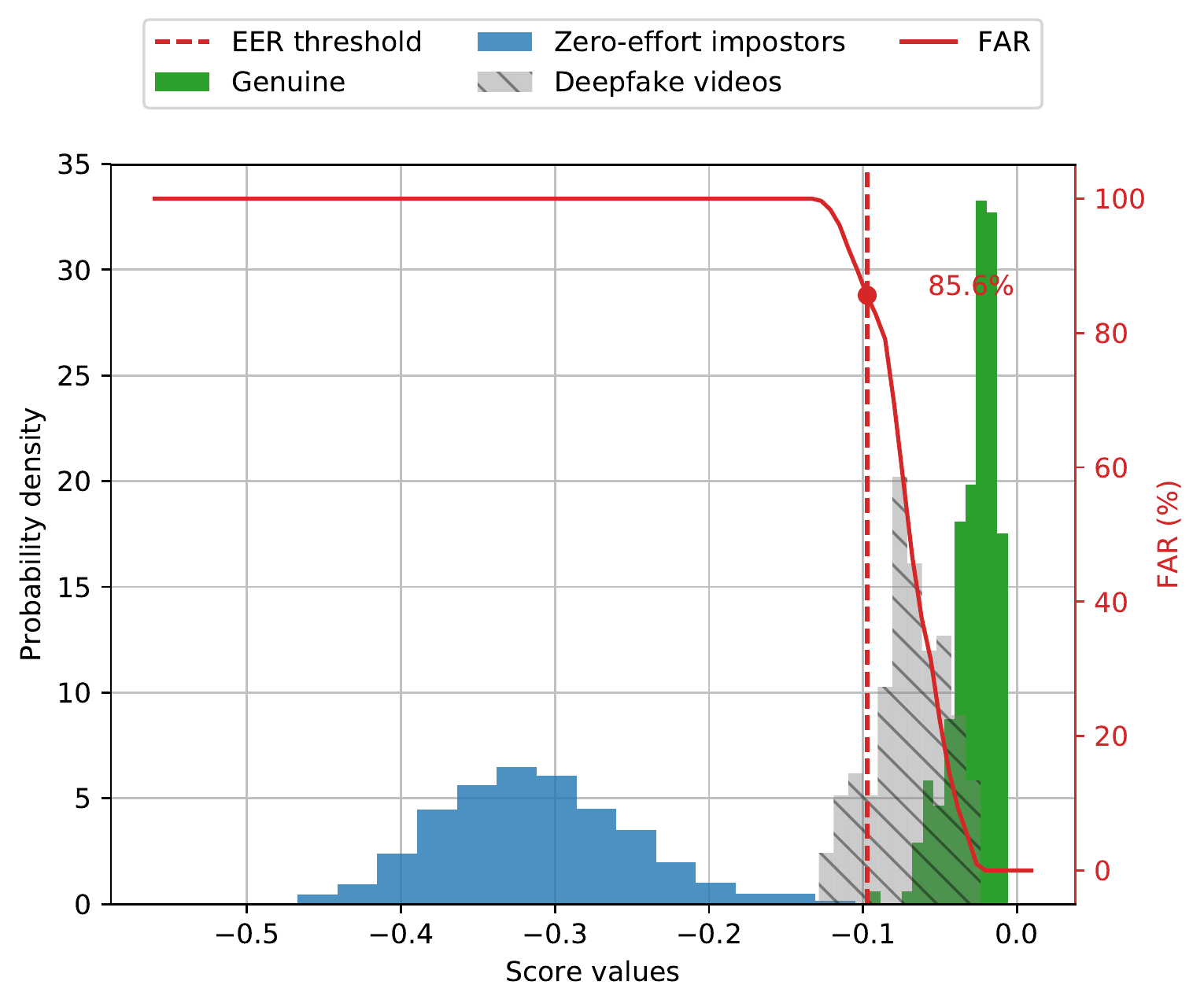}
  \label{fig:vulnvgg}}
\subfloat[FaceNet-based face recognition]{\includegraphics[width=\w]{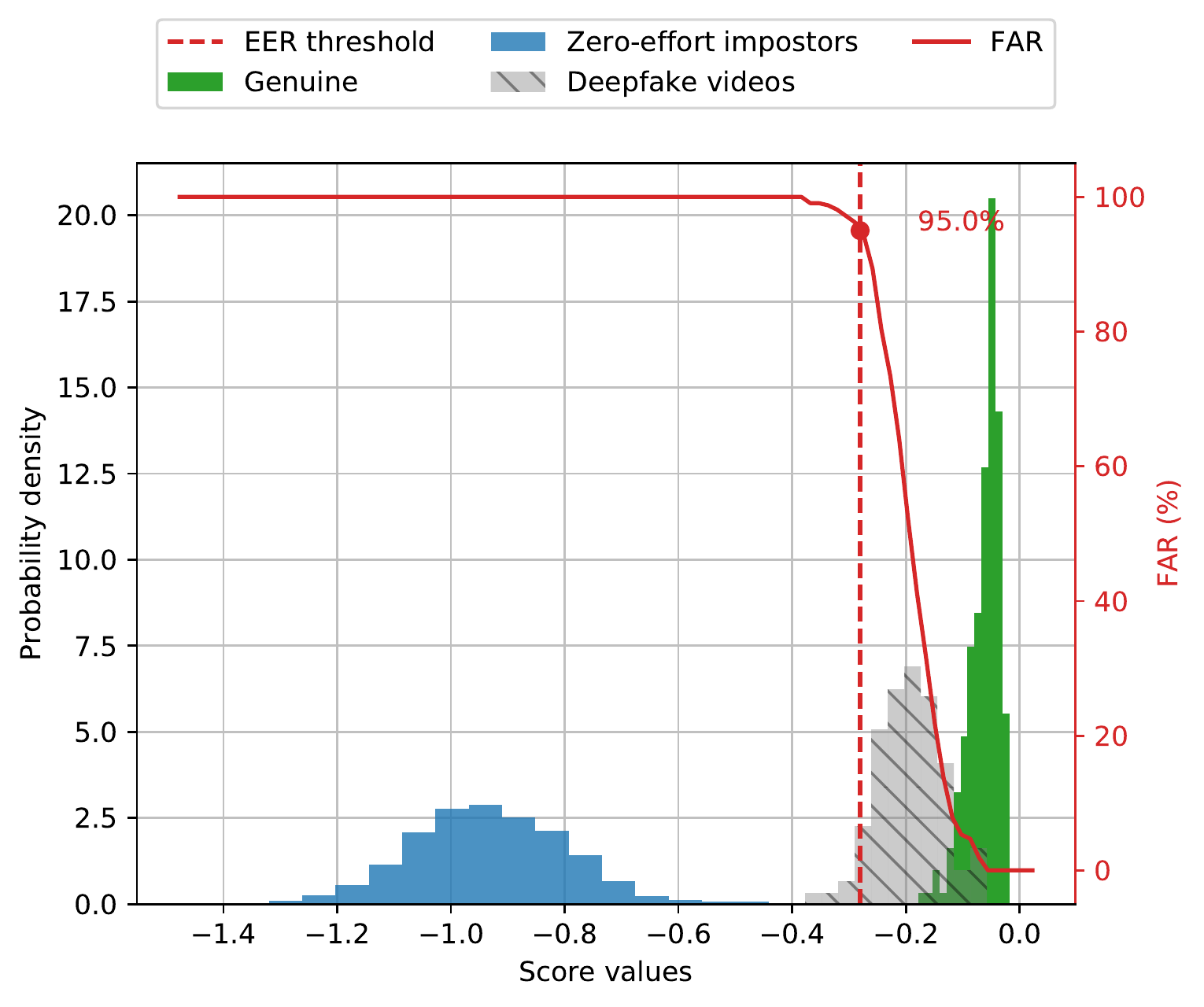}
  \label{fig:vulnfacenet}}
\caption{Histograms show the vulnerability of VGG and Facenet based face recognition to high quality deep morphs.}
\label{fig:vuln}
\end{figure*}

\subsection{Evaluation protocol}
\label{sec:protocol}

When evaluating vulnerability of face recognition, for the \textit{licit} scenario without the deep morph videos, we used the original VidTIMIT\footnotemark[5] videos for the $32$ subjects for which we have generated corresponding deep morph videos. In this scenario, we used $2$ videos of the subject for enrollment and the other $8$ videos as probes, for which we computed the verification scores. 


From the scores, for each possible threshold $\theta$, we computed commonly used metrics for evaluation of classification systems: false acceptance rate (FAR) and false reject rate (FRR).
Threshold at which these FAR and FRR are equal leads to an equal error rate (EER), which is commonly used as a single value metric of the system performance.

To evaluate vulnerability of face recognition, in \textit{tampered} scenario, we use deep morph videos ($10$ for each of $32$ subjects) as probes and compute the corresponding scores using the enrollment model from the \textit{licit} scenario. To understand if face recognition perceives deep morphs to be similar to the genuine original videos, we report the FAR metric computed using EER threshold $\theta$ from \textit{licit} scenario. If FAR value for deep morph videos is significantly higher than the one computed in \textit{licit} scenario, it means the face recognition system cannot distinguish synthetic videos from originals and is therefore vulnerable to deep morphs.

When evaluating deep morph detection, we consider it as a binary classification problem and evaluate the ability of detection approaches to distinguish original videos from deep morph videos. All videos in the dataset, including genuine and fake parts, were split into training (\textit{Train}) and evaluation (\textit{Test}) subsets. To avoid bias during training and testing, we arranged that the same subject would not appear in both sets. We did not introduce a development set, which is typically used to tune hyper parameters such as threshold, because the dataset is not large enough. Therefore, for deep morph detection system, we report the EER and the FRR (using the threshold when $FAR=10\%$) values on the \textit{Test} set.

%
%

\section{Vulnerability of face recognition}
\label{sec:vuln}

We used publicly available pre-trained VGG and Facenet architectures for face recognition. We used the \textit{fc7} and \textit{bottleneck} layers of these networks, respectively, as features and used cosine distance as a classifier. For a given test face, the confidence score of whether it belongs to a pre-enrolled model of a person is the cosine distance between the average feature vector, i.e., model, and the features vector of a test face. Both of these systems are state of the art recognition systems with VGG of $98.95\%$~\cite{Parkhi15} and Facenet of $99.63\%$~\cite{Schroff2015} accuracies on labeled faces in the wild (LFW) dataset.


We conducted the vulnerability analysis of VGG and Facenet-based face recognition systems on low quality (LQ) and high quality (HQ) face swaps in VidTIMIT\footnotemark[5] database. 
In a \textit{licit} scenario when only original videos are present, both systems performed very well, with EER value of $0.03\%$ for VGG and $0.00\%$ for Facenet-based system. Using the EER threshold from \textit{licit} scenario, we computed FAR value for the scenario when deep morph videos are used as probes. In this case, for VGG the FAR is $88.75\%$ on LQ deep morphs and $85.62\%$ on HQ deep morphs, and for Facenet the FAR is $94.38\%$ and $95.00\%$ on LQ and HQ deep morphs respectively. 
To illustrate this vulnerability, we plot the score histograms for high quality deep morph videos in Figure~\ref{fig:vuln}. The histograms show a considerable overlap between deep morph and genuine scores with clear separation from the zero-effort impostor scores (the probes from \textit{licit} scenario).

From the results, it is clear that both VGG and Facenet based systems cannot effectively distinguish GAN-generated synthetic faces from the original ones. The fact that more advanced Facenet system is more vulnerable is also consistent with the findings about presentation attacks~\cite{Mohammadi2018}. 

%
%
%
%
%


\section{Detection of deep morph videos}
\label{sec:detection}

We considered several baseline deep morph detection systems:

\vspace{-3pt}
\begin{itemize}
\item \textit{Pixels+PCA+LDA}: use raw faces as features with PCA-LDA classifier, with $99\%$ retained variance resulting in $446$ dimensions of transform matrix.
\vspace{-3pt}
\item \textit{IQM+PCA+LDA}: IQM features with PCA-LDA classifier with $95\%$ retained variance resulting in $2$ dimensions of transform matrix.
\vspace{-3pt}
\item \textit{IQM+SVM}: IQM features with SVM classifier, each video has an averaged score from $20$ frames.
\end{itemize}
\vspace{-3pt}

The systems based on image quality measures (IQM) are borrowed from the domain of presentation (including replay attacks) attack detection, where such systems have shown good performance~\cite{Galbally2014,Wen2015}. As IQM feature vector, we used $129$ measures of image quality, which include such measures like signal to noise ratio, specularity, bluriness, etc.,  by combining the features from~\cite{Galbally2014} and~\cite{Wen2015}. 

The results for all detection systems are presented in Table~\ref{tab:detection}. 
The results demonstrate that the IQM+SVM system has a reasonably high accuracy of detecting deep morph videos, although videos generated with HQ model pose a more serious challenge. It means that a more advanced techniques for face swapping will be even more challenging to detect.


\begin{table}[tb]
\footnotesize
\caption{Baseline detection systems for low (LQ) and high quality (HQ) deep morph videos. EER and FRR when FAR equal to 10\% are computed on Test set.}
\label{tab:detection}
\centering

\setlength\tabcolsep{1.5pt}
\def\arraystretch{1.3}%

\begin{tabular}{l|c|c|c}
\toprule

{\bf Database} & {\bf Detection system} & {\bf EER (\%)} & {\bf FRR@FAR10\% (\%)} \\ \midrule

& Pixels+PCA+LDA & 39.48 & 78.10 \\
LQ deep morph & IQM+PCA+LDA & 20.52 & 66.67\\
& IQM+SVM & 3.33 & 0.95 \\
\midrule
HQ deep morph & IQM+SVM & 8.97 & 9.05 \\
\bottomrule
\end{tabular}
\end{table}

%

\section{Conclusion}
In this paper, we demonstrated that state of the art VGG and Facenet-based face recognition algorithms are vulnerable to the deep morphed videos from DeepfaTIMIT database and fail to distinguish such videos from the original ones with up to $95.00\%$ equal error rate. We also evaluated several baseline detection algorithms and found that the techniques based on image quality measures with SVM classifier can detect HQ deep morph videos with $8.97$\% equal error rate. 

However, the continued advancements in development of GAN-generated faces will result in more challenging videos, which will be harder to detect by the existing algorithms. Therefore, new databases and new more generic detection methods need to be developed in the future. 


{\small
\bibliographystyle{ieee}
\bibliography{references_pavel}
}

\end{document}